\documentclass[runningheads]{llncs}
\pdfoutput=1
\usepackage{graphicx}
\usepackage{comment}
\usepackage{amsmath,amssymb} 
\usepackage{color}

\usepackage{epsfig}
\usepackage{tabularx}

\usepackage{subfig}
\usepackage[utf8]{inputenc}
\usepackage{tikz}
\usetikzlibrary{patterns}
\usepackage{pgfplots}
\pgfplotsset{compat=newest}
\usepgfplotslibrary{groupplots}
\usepgfplotslibrary{dateplot}
\usepgfplotslibrary{fillbetween}
\usepackage{booktabs}

\usepackage[breaklinks=true,bookmarks=false,hidelinks]{hyperref}
\usepackage{multirow}

\pgfmathdeclarefunction{poly}{0}{\pgfmathparse{tanh(2 / pi * 10 * x)}}
\pgfmathdeclarefunction{poly_mirrored}{0}{\pgfmathparse{tanh(2 / pi * 10 * (1 - x))}}

\renewcommand{\vec}[1]{\mathbf{#1}}

\newcommand{\etal}{\textit{et al}. }

\begin{document}
\pagestyle{headings}
\mainmatter
\def\ECCVSubNumber{23}  

\title{Evaluating Input Perturbation Methods for Interpreting CNNs and Saliency Map Comparison} 

\titlerunning{Evaluating Input Perturbation Methods for Interpreting CNNs}
%
\author{Lukas Brunke \and
Prateek Agrawal \and
Nikhil George}
%
\authorrunning{L. Brunke et al.}
%
\institute{Volkswagen Group of America, Belmont CA 94002, USA\\
\email{prateek.agrawal@vw.com}
}
\maketitle

\begin{abstract}
   Input perturbation methods occlude parts of an input to a function and measure the change in the function’s output. Recently, input perturbation methods have been applied to generate and evaluate saliency maps from convolutional neural networks. In practice, neutral baseline images are used for the occlusion, such that the baseline image’s impact on the classification probability is minimal. However, in this paper we show that arguably neutral baseline images still impact the generated saliency maps and their evaluation with input perturbations. We also demonstrate that many choices of hyperparameters lead to the divergence of saliency maps generated by input perturbations. We experimentally reveal inconsistencies among a selection of input perturbation methods and find that they lack robustness for generating saliency maps and for evaluating saliency maps as saliency metrics. 
\end{abstract}

\section{Introduction}

Understanding and interpreting convolutional neural networks' (CNN) predictions through saliency methods has become an active field of research in the recent years \cite{bach-plos15,grad_cam,Sundararajan2017AxiomaticAF,zeiler_fergus_2014}. Saliency methods create saliency maps, which highlight relevant parts of the input image for a classification task. 

\begin{figure}[htp!]
	\centering{
		\def\svgwidth{0.75\linewidth}
		\input{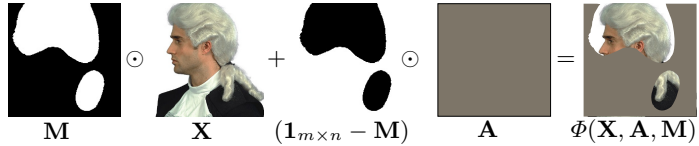}
		\caption{Masking an input image $\vec{X}$ with a baseline image $\vec{A}$. The input is elementwise multiplied with the mask $\vec{M}$, keeping all the parts of the input, where the mask has a value of $1$, and setting the parts of the input to zero, where the mask has a value of $0$. The baseline image is elementwise multiplied with the inverse mask ($\vec{1}_{m \times n} - \vec{M}$), which is the matrix of ones minus the mask, and added to the masked input, such that the parts, which were set to 0 for the input are set to the corresponding value of the baseline image. The displayed mask shows an intermediate mask from LeRF.}
		\label{fig:masking}
	}
\end{figure}

Input perturbation methods, which are also referred to as occlusion methods \cite{ancn17}, are one of the saliency methods for understanding CNNs. Input perturbation methods follow a simple principle: covering up or masking an important part of an input results in relevant information loss and should reduce the prediction score for a specific class. If an occluded part of the image results in an increase of the prediction score, then this part of the input will be negatively correlated with the target class. Furthermore, there might be parts of the input, which when covered, do not affect the prediction score. Occluding or masking an input is defined as substituting specific elements with a baseline image. The masking of an input image with a given baseline image and a mask is illustrated in Figure \ref{fig:masking}. In practice, neutral images are chosen as the baseline image, in such a way that the baseline image’s impact on the classification probability is minimal. Furthermore, input perturbation methods are also used to compare and evaluate saliency maps from different saliency methods for the same input image.
Input perturbation methods have two major advantages over other saliency methods like gradient-based methods. First, input perturbation methods allow the analysis of black-box models, whose weights and gradients are inaccessible. This is, for example, relevant in the automotive industry for validating black box models received from suppliers. Second, input perturbation methods' saliency maps are easily interpretable by humans unlike gradient-based methods, which often result in noisy or diffuse saliency maps \cite{NIPS2019_9167,ancn17}. In contrast to these advantages of input perturbation methods, a disadvantage is that they are typically more computationally intensive.

In this paper we experimentally evaluate the robustness of perturbation methods for image classification tasks against different parameters and baseline images. We pick three representative input perturbation methods to illustrate our findings: Randomized Input Sampling for Explanation (RISE) \cite{Petsiuk2018rise}, most relevant first (MoRF) \cite{samek_etal_2017}, and least relevant first (LeRF). However, our results also apply to other input perturbation methods. The goal is to determine the effects of changing the perturbation methods' parameters on the reliability of the perturbation method. We evaluate the generation of saliency maps with RISE by varying its parameters and selecting various baseline images, which are used to occlude an input. MoRF, and LeRF are metrics used to objectively compare different saliency maps for the same input by increasingly perturbing the original input \cite{samek_etal_2017}. We also vary the baseline images for MoRF and LeRF. The experiments reveal shortcomings regarding the robustness of input perturbation methods. 

The next section introduces the investigated perturbation methods. Section \ref{sec:related} gives an overview of related work. The material used for the experiments is presented in Section \ref{sec:material}. The experiments on RISE, MoRF, and LeRF are described in detail in Section \ref{sec:experiments} and subsequently discussed in Section \ref{sec:discussion}. We conclude our findings in Section \ref{sec:conclusion}.

\section{Input Perturbation Methods}
\label{sec:perturbation_methods}
Perturbation methods rely on a baseline image \cite{Sundararajan2017AxiomaticAF} to occlude parts of the original input, see Figure \ref{fig:masking}. The application of a baseline image $\vec{A} \in \mathbb{R}^{m \times n \times 3}$ with mask $\vec{M} \in \left[0, 1\right]^{m \times n}$ to an input image $\vec{X}\in \mathbb{R}^{m \times n \times 3}$ is defined in \cite{dabkowski_gal_2017,fong_iccv_2017} as:
\begin{equation}
    \Phi (\vec{X}, \vec{A}, \vec{M}) = \vec{M} \odot \vec{X} + (\vec{1}_{m \times n} - \vec{M}) \odot \vec{A}, 
\end{equation}
where $\odot$ denotes the Hadamard product and $\vec{1}_{m \times n}$ is an $\text{m} \times \text{n}$-dimensional matrix of ones. The mask can also be of a different dimension, depending on the scaling and cropping operations \cite{Petsiuk2018rise}.

No systematic pattern of how to choose a suitable baseline image seems identifiable in the literature. It has been argued to be a neutral input such that $\Tilde{f}(\vec{X}) \approx \vec{0}$, where $\Tilde{f}: \mathbb{R}^{m \times n \times 3} \xrightarrow{} \mathbb{R}^{K} $ is a CNN function and $K$ is the number of predicted classes \cite{Sundararajan2017AxiomaticAF}. The function $f = \Tilde{f} \circ g$ is the DNN function with preprocessing, where $g: \mathbb{R}^{m \times n \times 3} \xrightarrow{} \mathbb{R}^{m \times n \times 3}$ is the preprocessing function applied to the input $\vec{X}$ of a DNN. The black image $\vec{A} = \vec{0}_{m \times n \times 3}$, with the $\text{m} \times \text{n} \times \text{3}$-dimensional matrix of zeros $\vec{0}_{m \times n \times 3}$, is the baseline image used in \cite{Sundararajan2017AxiomaticAF}. Samek \etal applied baseline images from a uniform distribution, Dirichlet distribution, constant baseline images and blurred baseline images \cite{samek_etal_2017}. Petsiuk \etal use the baseline image for which $g\left(\vec{A}\right) = \vec{0}_{m \times n \times 3}$ \cite{Petsiuk2018rise}. We refer to this baseline image as the zero baseline image after preprocessing. 

\subsection{Saliency Map Generation: RISE}
Petsiuk \etal introduced RISE for creating saliency maps for interpreting CNNs for classification tasks \cite{Petsiuk2018rise}. RISE generates a saliency map by applying $N$ random masks to an input and creating a weighted sum of the $N$ output probabilities for the target class $c$ with the random masks. RISE uses a set of $N$ random masks $\mathcal{M}_N(p, w, h) = \{\vec{M}_1(p, w, h), ..., \allowbreak\vec{M}_N(p, w, h)\}$, where $\vec{M}_i(p, w, h) \in \{0, 1\}^{w \times h}, i \in \{1, ..., N\}$ and the probability $p \in \left[0,1\right]$ of each element of the mask to be equal to $1$ or in this context to be "on". The parameters $w$ and $h$ indicate the size of the low resolution mask before scaling and cropping. In the following $w = h$, since we only use square masks and input images. Each mask is scaled to the input image's size with a transformation $q : \{0, 1\}^{w \times h} \xrightarrow{} [0, 1]^{m \times n}$, which applies a bilinear transformation and a random cropping operation.  For a detailed description of the mask generation for RISE we refer the reader to the original paper \cite{Petsiuk2018rise}. The saliency map $S$ is then given by:
\begin{equation}
\vec{S} = \frac{1}{\mathbb{E}(\mathcal{M}_N) \cdot N} \sum_{i=1}^N \Tilde{f}^c(q(\vec{M}_i) \odot g(\vec{X})),
\end{equation}
where  $\Tilde{f}^c$ are the outputs for class $c$ for the DNN function without preprocessing. Directly applying the mask to the preprocessed input is equal to implicitly using the zero after preprocessing baseline image, therefore, using a constant baseline image for the saliency map generation: 
\begin{equation}
\Tilde{f}^c(q(\vec{M}_i) \odot g(\vec{X})) = f^c(\Phi(\vec{X}, g^{-1}(\vec{0}_{m \times n \times 3}), q(\vec{M}_i))
\label{eq:zero_after_preprocessing}
\end{equation}
with the inverse preprocessing function $g^{-1}$ and $f^c$, which is the output for class $c$ for the DNN function with preprocessing.

\subsection{Saliency Map Comparison: MoRF and LeRF}
The MoRF and LeRF metrics quantitatively evaluate the quality of saliency maps \cite{samek_etal_2017}. This is relevant for comparing and assessing different saliency methods like RISE \cite{Petsiuk2018rise}, Grad-CAM \cite{grad_cam}, and integrated gradients \cite{Sundararajan2017AxiomaticAF}. MoRF and LeRF measure the effect of occluding pixels from the input image on the target class' output probabilities. MoRF and LeRF replace elements from the input with the corresponding elements from a baseline image. Given a saliency map for an input image as a ranking of importance for individual pixels, MoRF replaces pixels in the input image in decreasing order of the importance ($=$ most relevant first), whereas LeRF replaces pixels in the input image in increasing order of importance ($=$ least relevant first). Replacing pixels in the input image with the corresponding pixels from the baseline image results in a change of the output probabilities. Recording the output probabilities as  further elements are occluded results in a curve over the relative number of occluded elements $\alpha$. The number of additionally occluded pixels $r$ in each step is variable. Here, all presented experiments are run with $r = 1$, occluding only one additional pixel in each step.

MoRF iteratively replaces elements with decreasing importance according to a given ordering by a saliency map. A greater area over the curve (AOC) for MoRF is desirable and suggests a superior saliency map. On the other hand, LeRF iteratively replaces elements with increasing importance and its score is given by the area under the curve (AUC). In this paper we limit the experiments to the MoRF and LeRF metrics, because the MoRF and LeRF metrics can be transformed into the insertion and deletion metrics, which are used in \cite{fong_iccv_2017,Petsiuk2018rise}.

\section{Related Work}
\label{sec:related}
There has been a limited effort to determine the robustness of input perturbation methods against varying parameters and baseline images. Ancona \etal \cite{ancn17} have tested different square sizes for masks for the sliding window method \cite{zeiler_fergus_2014}. Increased square sizes exhibited a reduction of details provided in the saliency maps. This inspired our experiments on varying the $w$ and $p$ parameters for RISE, which affect the shape and size of the random masks. Similarly to Ancona \etal \cite{ancn17}, we find that less fine grained saliency maps are created by less fine grained masks. This corresponds to small values for $w$ in the case of RISE and bigger values for $n$ in the case of Occlusion-$n$. 

Samek \etal \cite{samek_etal_2017} have investigated various baseline images for the MoRF and LeRF metrics. They determined that the uniform distribution baseline image gives the best MoRF and LeRF scores when averaged over the whole dataset. Samek \etal applied MoRF and LeRF with $r = 9 \times 9$ non-overlapping regions and only perturbed up to $15.7\%$ of the input image. Furthermore, they evaluated the blurred baseline image with $\sigma = 3$, which is arguably small and still contains information from the original input. 

Fong and Vedaldi \cite{fong_iccv_2017} generate saliency maps by optimizing perturbation masks, which minimize or maximize the output probabilities. They find that different baseline images yield different saliency maps, which highlights the dependency on the choice of baseline image. Dabkowski and Gal \cite{dabkowski_gal_2017} note similar findings and therefore use a set of baseline images to circumvent the dependence on a single baseline image. Petsiuk \etal \cite{Petsiuk2018rise} determined through qualitative reasoning for MoRF and through quantitative analysis for LeRF, that a constant gray baseline image and a blurred baseline image work best for their evaluation using MoRF (= deletion) and LeRF (= insertion), respectively. However, our quantitative analysis suggests that this is not the case.

Recently, \cite{sanity_checks_for_metrics} have investigated the reliability of saliency metrics like MoRF and LeRF. Using two different baseline images, specifically the constant dataset mean baseline image and a uniform noise baseline image, they show that MoRF and LeRF are dependent on the choice of the baseline image. However, since they used a random baseline image for comparison, the baseline image changes for each run of MoRF and LeRF, which lacks consistency and could affect their findings and yield issues regarding reproducibility. In contrast, we use a set of constant baseline images (with the exception of the blurred baseline image). Additionally, they only analyzed the scores for MoRF and LeRF for 100 steps, whereas we run both metrics for all steps.

\section{Material}
\label{sec:material}
In this work we are exclusively considering CNNs for image classification. Specifically we are using the pre-trained CNN ResNet-50 \cite{He_2016_CVPR} from Keras \cite{chollet2015keras} in TensorFlow \cite{tensorflow2015-whitepaper}. We process the input in the same way as described by \cite{He_2016_CVPR}, such that the input shape is $224 \times 224 \times 3$. Then we apply a mean shift, which is the preprocessing function used for the pretrained network.

All experiments are run with the validation set from ImageNet \cite{imagenet_cvpr09}, which we refer to as dataset $\mathcal{D}_{\mathrm{val}}$. Saliency maps are visualized by normalizing the resulting maps in the range $[0, 1]$ and applying OpenCV's perceptually uniform sequential inferno colormap \cite{opencv_library}. This colormap displays parts supporting the prediction with lighter colors and parts opposing the prediction with darker colors.

\section{Experiments}
\label{sec:experiments}
This section presents experiments on the robustness of input perturbation methods against different parameter settings and different baseline images. First, we investigate how saliency maps change under varying parameters for the mask generation with RISE. Second, we analyze the impact of using different supposedly uninformative baseline images on the robustness of RISE, MoRF, and LeRF. 

\subsection{Experiments on RISE}
Petsiuk \etal apply RISE with $N = 8000$ for ResNet-50, $h = w = 7$, $p = 0.5$ and the zero baseline image after preprocessing \cite{Petsiuk2018rise}. The choice of these parameters is not elaborated on in \cite{Petsiuk2018rise}. Therefore, this section determines the effect of varying the parameters on the convergence of a saliency map and the subjective quality of the saliency map. Throughout this section, saliency maps are generated for their maximum activated class $c_\mathrm{max} = \arg \max_c f^c(\vec{x})$. In contrast to Petsiuk \etal \cite{Petsiuk2018rise}, we applied $N_{\mathrm{max}} = 32768$ masks to evaluate the convergence. Unless otherwise stated the experiments are executed with $h=w=7$, $p=0.5$, $N = N_{\mathrm{max}}$, and the zero baseline image after preprocessing.

\begin{figure}[htp!]
\begin{center}
    \subfloat[$\vec{X}$]{{\includegraphics[width=0.14\linewidth]{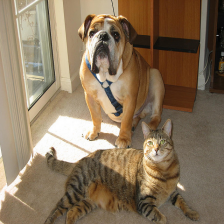}}}%
	\hfill
    \subfloat[$\vec{S}_1$]{{\includegraphics[width=0.14\linewidth]{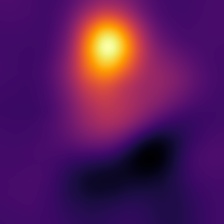}}}%
	\hfill
    \subfloat[$\vec{S}_2$]{{\includegraphics[width=0.14\linewidth]{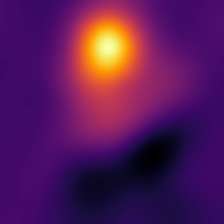}}}%
	\hfill
    \subfloat[$\vec{S}_3$]{{\includegraphics[width=0.14\linewidth]{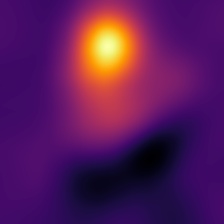}}}%
	\hfill
    \subfloat[Convergence plot]{\label{fig:plot_convergence_catdog}\raisebox{-11mm}{\input{figures/convergence/catdog/convergence.tex}}}%
\end{center}
   \caption{Convergence for RISE for the class \texttt{bull\_mastiff}.}
\label{fig:rise_conv_catdog}
\end{figure}

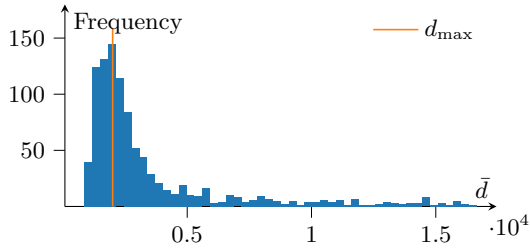
\begin{figure}
	\begin{center}
\begin{tikzpicture}

\definecolor{color0}{rgb}{0.12156862745098,0.466666666666667,0.705882352941177}
\definecolor{color1}{rgb}{1,0.498039215686275,0.0549019607843137}

\begin{axis}[
width=0.6\textwidth,
height=0.35\textwidth,
axis lines=middle,
legend cell align={left},
legend style={fill opacity=0.8, draw opacity=1, text opacity=1, draw=white!80.0!black},
every x tick scale label/.style={
	at={(xticklabel* cs:1.03,0cm)},
	anchor=near xticklabel
},
tick align=outside,
tick pos=both,
x grid style={white!69.01960784313725!black},
xlabel={$\bar{d}$},
xmin=79.428001595767, xmax=17445.2259274943,
xtick style={color=black},
y grid style={white!69.01960784313725!black},
ylabel={Frequency},
ymin=0, ymax=180,
ytick style={color=black},
legend style={draw=none, fill=none}
]
\draw[fill=color0,draw opacity=0] (axis cs:868.782452772974,0) rectangle (axis cs:1184.52423324386,40);
\draw[fill=color0,draw opacity=0] (axis cs:1184.52423324386,0) rectangle (axis cs:1500.26601371474,124);
\draw[fill=color0,draw opacity=0] (axis cs:1500.26601371474,0) rectangle (axis cs:1816.00779418562,131);
\draw[fill=color0,draw opacity=0] (axis cs:1816.00779418562,0) rectangle (axis cs:2131.7495746565,145);
\draw[fill=color0,draw opacity=0] (axis cs:2131.74957465651,0) rectangle (axis cs:2447.49135512739,114);
\draw[fill=color0,draw opacity=0] (axis cs:2447.49135512739,0) rectangle (axis cs:2763.23313559827,84);
\draw[fill=color0,draw opacity=0] (axis cs:2763.23313559827,0) rectangle (axis cs:3078.97491606915,52);
\draw[fill=color0,draw opacity=0] (axis cs:3078.97491606915,0) rectangle (axis cs:3394.71669654004,44);
\draw[fill=color0,draw opacity=0] (axis cs:3394.71669654004,0) rectangle (axis cs:3710.45847701092,29);
\draw[fill=color0,draw opacity=0] (axis cs:3710.45847701092,0) rectangle (axis cs:4026.2002574818,21);
\draw[fill=color0,draw opacity=0] (axis cs:4026.2002574818,0) rectangle (axis cs:4341.94203795268,15);
\draw[fill=color0,draw opacity=0] (axis cs:4341.94203795268,0) rectangle (axis cs:4657.68381842357,11);
\draw[fill=color0,draw opacity=0] (axis cs:4657.68381842357,0) rectangle (axis cs:4973.42559889445,19);
\draw[fill=color0,draw opacity=0] (axis cs:4973.42559889445,0) rectangle (axis cs:5289.16737936533,10);
\draw[fill=color0,draw opacity=0] (axis cs:5289.16737936533,0) rectangle (axis cs:5604.90915983622,9);
\draw[fill=color0,draw opacity=0] (axis cs:5604.90915983622,0) rectangle (axis cs:5920.6509403071,16);
\draw[fill=color0,draw opacity=0] (axis cs:5920.6509403071,0) rectangle (axis cs:6236.39272077798,3);
\draw[fill=color0,draw opacity=0] (axis cs:6236.39272077798,0) rectangle (axis cs:6552.13450124886,4);
\draw[fill=color0,draw opacity=0] (axis cs:6552.13450124886,0) rectangle (axis cs:6867.87628171975,10);
\draw[fill=color0,draw opacity=0] (axis cs:6867.87628171975,0) rectangle (axis cs:7183.61806219063,8);
\draw[fill=color0,draw opacity=0] (axis cs:7183.61806219063,0) rectangle (axis cs:7499.35984266151,4);
\draw[fill=color0,draw opacity=0] (axis cs:7499.35984266151,0) rectangle (axis cs:7815.10162313239,6);
\draw[fill=color0,draw opacity=0] (axis cs:7815.10162313239,0) rectangle (axis cs:8130.84340360328,9);
\draw[fill=color0,draw opacity=0] (axis cs:8130.84340360328,0) rectangle (axis cs:8446.58518407416,7);
\draw[fill=color0,draw opacity=0] (axis cs:8446.58518407416,0) rectangle (axis cs:8762.32696454504,5);
\draw[fill=color0,draw opacity=0] (axis cs:8762.32696454504,0) rectangle (axis cs:9078.06874501593,2);
\draw[fill=color0,draw opacity=0] (axis cs:9078.06874501593,0) rectangle (axis cs:9393.81052548681,5);
\draw[fill=color0,draw opacity=0] (axis cs:9393.81052548681,0) rectangle (axis cs:9709.55230595769,1);
\draw[fill=color0,draw opacity=0] (axis cs:9709.55230595769,0) rectangle (axis cs:10025.2940864286,4);
\draw[fill=color0,draw opacity=0] (axis cs:10025.2940864286,0) rectangle (axis cs:10341.0358668995,4);
\draw[fill=color0,draw opacity=0] (axis cs:10341.0358668995,0) rectangle (axis cs:10656.7776473703,6);
\draw[fill=color0,draw opacity=0] (axis cs:10656.7776473703,0) rectangle (axis cs:10972.5194278412,4);
\draw[fill=color0,draw opacity=0] (axis cs:10972.5194278412,0) rectangle (axis cs:11288.2612083121,6);
\draw[fill=color0,draw opacity=0] (axis cs:11288.2612083121,0) rectangle (axis cs:11604.002988783,1);
\draw[fill=color0,draw opacity=0] (axis cs:11604.002988783,0) rectangle (axis cs:11919.7447692539,7);
\draw[fill=color0,draw opacity=0] (axis cs:11919.7447692539,0) rectangle (axis cs:12235.4865497248,1);
\draw[fill=color0,draw opacity=0] (axis cs:12235.4865497248,0) rectangle (axis cs:12551.2283301956,1);
\draw[fill=color0,draw opacity=0] (axis cs:12551.2283301956,0) rectangle (axis cs:12866.9701106665,2);
\draw[fill=color0,draw opacity=0] (axis cs:12866.9701106665,0) rectangle (axis cs:13182.7118911374,4);
\draw[fill=color0,draw opacity=0] (axis cs:13182.7118911374,0) rectangle (axis cs:13498.4536716083,3);
\draw[fill=color0,draw opacity=0] (axis cs:13498.4536716083,0) rectangle (axis cs:13814.1954520792,2);
\draw[fill=color0,draw opacity=0] (axis cs:13814.1954520792,0) rectangle (axis cs:14129.93723255,3);
\draw[fill=color0,draw opacity=0] (axis cs:14129.93723255,0) rectangle (axis cs:14445.6790130209,3);
\draw[fill=color0,draw opacity=0] (axis cs:14445.6790130209,0) rectangle (axis cs:14761.4207934918,8);
\draw[fill=color0,draw opacity=0] (axis cs:14761.4207934918,0) rectangle (axis cs:15077.1625739627,1);
\draw[fill=color0,draw opacity=0] (axis cs:15077.1625739627,0) rectangle (axis cs:15392.9043544336,3);
\draw[fill=color0,draw opacity=0] (axis cs:15392.9043544336,0) rectangle (axis cs:15708.6461349045,1);
\draw[fill=color0,draw opacity=0] (axis cs:15708.6461349045,0) rectangle (axis cs:16024.3879153753,5);
\draw[fill=color0,draw opacity=0] (axis cs:16024.3879153753,0) rectangle (axis cs:16340.1296958462,2);
\draw[fill=color0,draw opacity=0] (axis cs:16340.1296958462,0) rectangle (axis cs:16655.8714763171,1);
\addplot [semithick, color1]
table {%
2000 0
2000 1
2000 2
2000 3
2000 4
2000 5
2000 6
2000 7
2000 8
2000 9
2000 10
2000 11
2000 12
2000 13
2000 14
2000 15
2000 16
2000 17
2000 18
2000 19
2000 20
2000 21
2000 22
2000 23
2000 24
2000 25
2000 26
2000 27
2000 28
2000 29
2000 30
2000 31
2000 32
2000 33
2000 34
2000 35
2000 36
2000 37
2000 38
2000 39
2000 40
2000 41
2000 42
2000 43
2000 44
2000 45
2000 46
2000 47
2000 48
2000 49
2000 50
2000 51
2000 52
2000 53
2000 54
2000 55
2000 56
2000 57
2000 58
2000 59
2000 60
2000 61
2000 62
2000 63
2000 64
2000 65
2000 66
2000 67
2000 68
2000 69
2000 70
2000 71
2000 72
2000 73
2000 74
2000 75
2000 76
2000 77
2000 78
2000 79
2000 80
2000 81
2000 82
2000 83
2000 84
2000 85
2000 86
2000 87
2000 88
2000 89
2000 90
2000 91
2000 92
2000 93
2000 94
2000 95
2000 96
2000 97
2000 98
2000 99
2000 100
2000 101
2000 102
2000 103
2000 104
2000 105
2000 106
2000 107
2000 108
2000 109
2000 110
2000 111
2000 112
2000 113
2000 114
2000 115
2000 116
2000 117
2000 118
2000 119
2000 120
2000 121
2000 122
2000 123
2000 124
2000 125
2000 126
2000 127
2000 128
2000 129
2000 130
2000 131
2000 132
2000 133
2000 134
2000 135
2000 136
2000 137
2000 138
2000 139
2000 140
2000 141
2000 142
2000 143
2000 144
2000 145
2000 146
2000 147
2000 148
2000 149
2000 150
2000 151
2000 152
2000 153
2000 154
2000 155
2000 156
2000 157
2000 158
2000 159
};
\addlegendentry{$d_{\mathrm{max}}$}
\end{axis}

\end{tikzpicture}
	\end{center}
	\caption{Selection of $d_{\mathrm{max}} = 2000$.}
	\label{fig:histogram}
\end{figure}

First, the number of masks needed for convergence of a single saliency map is determined. In order to check the convergence RISE is run three times, with three independent random sets of masks $\mathcal{M}^l_{N_{\mathrm{max}}}$, where $l \in \{1, 2, 3\}$ and $N_{\mathrm{max}} = 32768$. Figure \ref{fig:rise_conv_catdog} displays the resulting saliency maps for an input $\vec{X}$ after applying $N_{\mathrm{max}}$ masks. Figure \ref{fig:rise_conv_catdog} shows that the saliency maps from multiple runs have converged, since they are nearly indistinguishable from each other. 
Since the visual assessment of convergence for saliency maps is not efficient, the goal is to determine convergence quantitatively. We define a function $d$ that calculates the $\mathrm{L}^2$-distance between two saliency maps  $S_i$ and $S_j$, where $i \neq j, i \in \{1, 2, 3\}$ and $j \in \{1, 2, 3\}$,  from different runs and records the $\mathrm{L}^2$-distance for every incremental saliency map. In practice, we only consider saliency maps with a maximum distance between saliency maps from independent runs smaller than $d_{\mathrm{max}} = 2000$, which is represented by the dashed horizontal line in Figure \ref{fig:plot_convergence_catdog}. We chose this threshold, because it indicates the maximum in the histogram in Figure \ref{fig:histogram}. The histogram shows the $\mathrm{L}^2$-distances for RISE saliency maps from 1000 randomly sampled input images from ImageNet. Manual inspection confirmed that the threshold also yields subjectively good results. Note, that for this input image the use of the suggested number of masks $N = 8000$, which is represented by the dotted vertical line in Figure \ref{fig:plot_convergence_catdog}, results in $d > d_\mathrm{max}$ for some combinations of $i$ and $j$.

\begin{figure}[htp!]
\begin{center}
    \subfloat[$\vec{X}$]{{\includegraphics[width=0.14\linewidth]{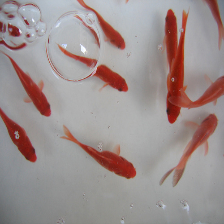}}}%
	\hfill
    \subfloat[$\vec{S}_1$]{{\includegraphics[width=0.14\linewidth]{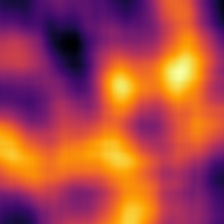}}}%
	\hfill
    \subfloat[$\vec{S}_2$]{{\includegraphics[width=0.14\linewidth]{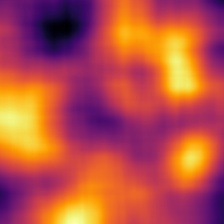}}}%
	\hfill
    \subfloat[$\vec{S}_3$]{{\includegraphics[width=0.14\linewidth]{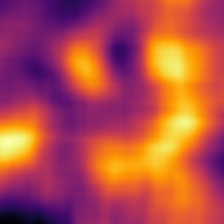}}}%
	\hfill
   \subfloat[Convergence plot]{\label{fig:plot_convergence_goldfish}\raisebox{-11mm}{\input{figures/convergence/goldfish/convergence.tex}}}\\%
	\subfloat[$\vec{X}$]{{\includegraphics[width=0.14\linewidth]{images/inputs_resized/goldfish_0.png}}}%
	\hfill
	\subfloat[$\vec{S}_1$]{{\includegraphics[width=0.14\linewidth]{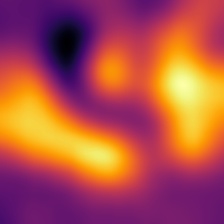}}}%
	\hfill
	\subfloat[$\vec{S}_2$]{{\includegraphics[width=0.14\linewidth]{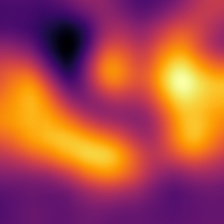}}}%
	\hfill
	\subfloat[$\vec{S}_3$]{{\includegraphics[width=0.14\linewidth]{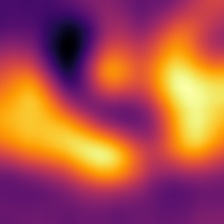}}}%
	\hfill
	\subfloat[Convergence plot]{\label{fig:plot_convergence}\raisebox{-11mm}{\input{figures/convergence/goldfish_10/convergence.tex}}}%
\end{center}
   \caption{Convergence for RISE with $p = 0.5$ at the top and $p = 0.1$ at the bottom for the class \texttt{goldfish}. The choice of $p=0.5$ does not lead to convergence, while $p=0.1$ converges. We found that only 389 of 1000 randomly sampled input images led to convergence with $p = 0.5$.}
\label{fig:rise_conv_goldfish}
\end{figure}

\begin{figure}[htp!]
	\begin{center}
		\subfloat[$\vec{X}$]{{\includegraphics[width=0.14\linewidth]{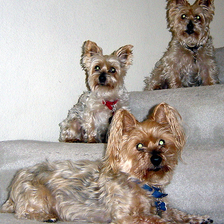}}}%
		\hfill
		\subfloat[$w = 3$]{{\includegraphics[width=0.14\linewidth]{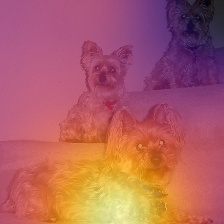}}}%
		\hfill
		\subfloat[$w = 5$]{{\includegraphics[width=0.14\linewidth]{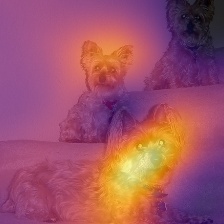}}}%
		\hfill
		\subfloat[$w = 7$]{{\includegraphics[width=0.14\linewidth]{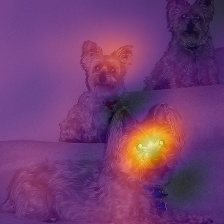}}}%
		\hfill
		\subfloat[Convergence plot]{\label{fig:w_799_plot}\raisebox{-8mm}{
\begin{tikzpicture}

\definecolor{color0}{rgb}{0.12156862745098,0.466666666666667,0.705882352941177}

\begin{axis}[
xlabel={$w$},
xmin=2.8, xmax=7.2,
ymin=0, ymax=2500,
ytick style={color=black},
legend cell align={left},
legend style={at={(0.8,0.9)}, anchor=north},
axis lines=middle,
xlabel near ticks,
xtick style={color=black},
ylabel={$\bar{d}$},
ytick style={color=black},
y tick label style={
	/pgf/number format/.cd,
	fixed,
	fixed zerofill,
	precision=0,
	/tikz/.cd
},
x tick label style={
	/pgf/number format/.cd,
	fixed,
	fixed zerofill,
	precision=0,
	/tikz/.cd
},
legend style={draw=none},
height=0.3\textwidth,
width=0.4\linewidth
]
\addplot [semithick, color0, mark=*, mark size=3, mark options={solid}, only marks]
table {%
3 645.032370285416
4 872.607034851031
5 1892.78407425685
7 1855.25612357373
};
\end{axis}

\end{tikzpicture}}}\\%
		\subfloat[$\vec{X}$]{{\includegraphics[width=0.14\linewidth]{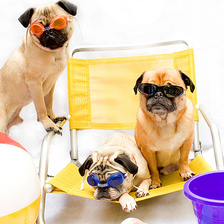}}}%
		\hfill
		\subfloat[$w = 3$]{{\includegraphics[width=0.14\linewidth]{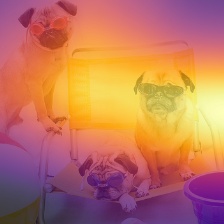}}}%
		\hfill
		\subfloat[$w = 5$]{{\includegraphics[width=0.14\linewidth]{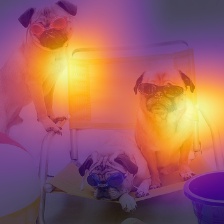}}}%
		\hfill
		\subfloat[$w = 7$]{{\includegraphics[width=0.14\linewidth]{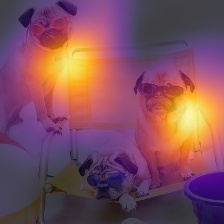}}}%
		\hfill
		\subfloat[Convergence plot]{\label{fig:plot_convergence}\raisebox{-8mm}{
\begin{tikzpicture}

\definecolor{color0}{rgb}{0.12156862745098,0.466666666666667,0.705882352941177}

\begin{axis}[
xlabel={$w$},
xmin=2.8, xmax=7.2,
ymin=0, ymax=2500,
ytick style={color=black},
legend cell align={left},
legend style={at={(0.8,0.9)}, anchor=north},
axis lines=middle,
xlabel near ticks,
xtick style={color=black},
ylabel={$\bar{d}$},
ytick style={color=black},
y tick label style={
	/pgf/number format/.cd,
	fixed,
	fixed zerofill,
	precision=0,
	/tikz/.cd
},
x tick label style={
	/pgf/number format/.cd,
	fixed,
	fixed zerofill,
	precision=0,
	/tikz/.cd
},
legend style={draw=none},
height=0.3\textwidth,
width=0.4\linewidth
]
\addplot [semithick, color0, mark=*, mark size=3, mark options={solid}, only marks]
table {%
3 1898.64906812017
4 1685.01236461537
5 1666.12310787207
7 1421.7001348381
};
\end{axis}

\end{tikzpicture}}}%
	\end{center}
	\caption{Convergence for RISE with varying $w$ for the classes \texttt{Yorkshire terrier} and \texttt{pug}, respectively. The choice of $w$ can either reduce or increase $\bar{d}$.}
	\label{fig:rise_w_799}
\end{figure}

The input image and saliency maps in Figure \ref{fig:rise_conv_goldfish} show an example, where the standard parameters for $w$ and $p$ with $N_{\mathrm{max}} = 32768$ masks do not lead to the same saliency map for the three independent runs. The graph in Figure \ref{fig:plot_convergence_goldfish} shows that the $\mathrm{L}^2$-distance between the saliency maps in Figure \ref{fig:rise_conv_goldfish} never get close to the threshold $d_{\mathrm{max}}$, and therefore do not converge. Doing the same investigation on a random subset of 1000 images from the validation set from ImageNet, yields only 389 converged saliency maps. 
In the following we investigate the variation of the $p$ and $w$ parameters. We observed, that the example in Figure \ref{fig:rise_conv_goldfish}a-e converged when setting $p = 0.1$, which is shown in Figure \ref{fig:rise_conv_goldfish}f-j. Note, that the threshold $d_{\mathrm{max}}$ is only reached when $N$ approaches $N_{\mathrm{max}}$.
Applying different values for $w$, while $p$ is constant, results in distinct saliency maps, which also impacts the average $\mathrm{L}^2$-distance $\bar{d}$ between saliency maps from different runs at $N_{\mathrm{max}}$. Figure \ref{fig:rise_w_799} shows the effect of selecting different $w\in \{3, 4, 5, 7\}$. In the two examples, an increasing value for $w$ leads to more defined distinct local optima in the saliency maps, which focus on the individual objects of the target class. While $\bar{d}$ increases with increasing $w$ for the example in Figure \ref{fig:rise_w_799}a-e, $\bar{d}$ decreases with increasing $w$ for the example in Figure \ref{fig:rise_w_799}f-j.

\begin{figure}[htp!]
\begin{center}
    \subfloat[$(5,0.1)$]{{\includegraphics[width=0.14\linewidth]{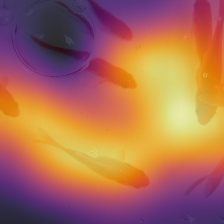}}}%
	\quad
    \subfloat[$(7,0.1)$]{{\includegraphics[width=0.14\linewidth]{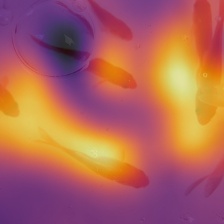}}}%
	\quad
    \subfloat[$(9,0.1)$]{{\includegraphics[width=0.14\linewidth]{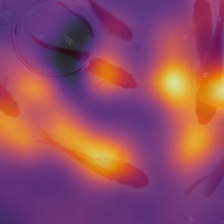}}}%
	\quad
    \subfloat[$(11,0.1)$]{{\includegraphics[width=0.14\linewidth]{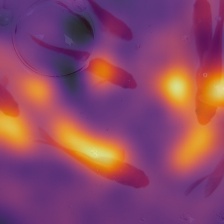}}}%

    \subfloat[$(5,0.3)$]{{\includegraphics[width=0.14\linewidth]{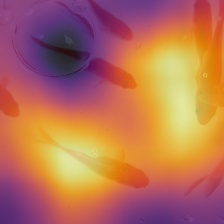}}}%
	\quad
    \subfloat[$(7,0.3)$]{{\includegraphics[width=0.14\linewidth]{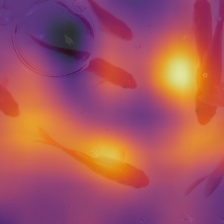}}}%
	\quad
     \subfloat[$(9,0.3)$]{{\includegraphics[width=0.14\linewidth]{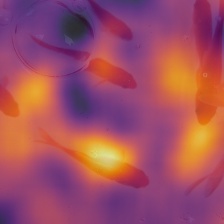}}}%
	\quad
    \subfloat[$(11,0.3)$]{{\includegraphics[width=0.14\linewidth]{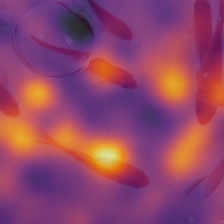}}}%
\end{center}
   \caption{RISE with different values for $(w,p)$ for the class \texttt{goldfish}. The choice of different hyperparameters yield different saliency maps.}
\label{fig:rise_convergence}
\end{figure}

Subsequently, the impact of varying $p$ and $w=h$ jointly is investigated. We analyzed the variation of $p \in \{0.1, 0,3, 0.5, 0.7, 0.9\}$ and $h = w \in \{5, 7, 9, 11\}$, where each set of parameters is run three times to determine convergence. In Figure \ref{fig:rise_convergence} we only report the saliency maps, which converged. Saliency maps with $p \in \{0.5, 0.7, 0.9\}$ did not converge for the input image from Figure \ref{fig:rise_conv_goldfish}. For some combinations the maximum number of masks $N_{\mathrm{max}} = 32768$ might not be sufficient for the saliency maps to converge. However, we did not run any experiments with $N > N_{\mathrm{max}}$, since the application of more masks increases the number of calculations, which leads to computational infeasibility in practical settings. 

\begin{figure}[htp!]
	\begin{center}
		\subfloat[$\vec{X}_1$]{{\includegraphics[width=0.14\linewidth]{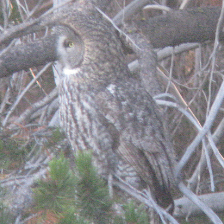}}}%
		\quad
		\subfloat[$\vec{X}_2$]{{\includegraphics[width=0.14\linewidth]{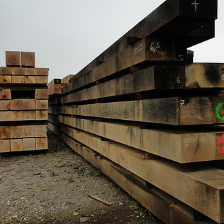}}}%
		\quad
		\subfloat[$\vec{X}_3$]{{\includegraphics[width=0.14\linewidth]{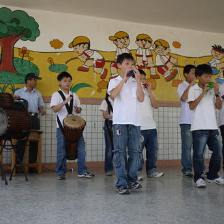}}}%
		\quad
		\subfloat[$\vec{X}_4$]{{\includegraphics[width=0.14\linewidth]{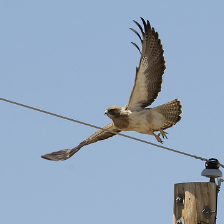}}}%
		
		\subfloat[$\vec{A}_{\mathrm{inv}}$]{{\includegraphics[width=0.14\linewidth]{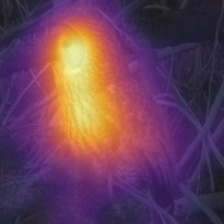}}}%
		\quad
		\subfloat[$\vec{A}_{255}$]{{\includegraphics[width=0.14\linewidth]{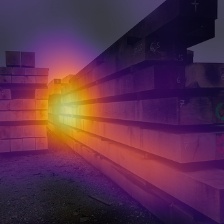}}}%
		\quad
		\subfloat[$\vec{A}_{127}$]{{\label{fig:rise_different_baselines_gray_people}\includegraphics[width=0.14\linewidth]{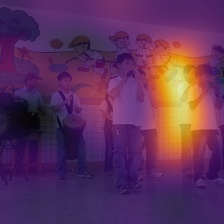}}}%
		\quad
		\subfloat[$\vec{A}_{255}$]{{\label{fig:rise_different_baselines_white_bird}\includegraphics[width=0.14\linewidth]{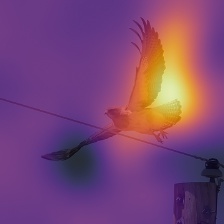}}}%
		
		\subfloat[$\vec{A}_{\sigma}$]{{\includegraphics[width=0.14\linewidth]{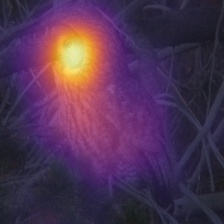}}}%
		\quad
		\subfloat[$\vec{A}_{0}$]{{\includegraphics[width=0.14\linewidth]{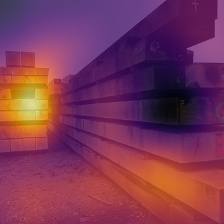}}}%
		\quad
		\subfloat[$\vec{A}_{0}$]{{\label{fig:rise_different_baselines_black_people}\includegraphics[width=0.14\linewidth]{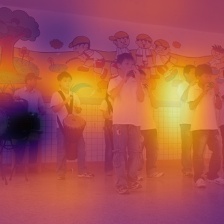}}}%
		\quad
		\subfloat[$\vec{A}_{0}$]{{\label{fig:rise_different_baselines_black_bird}\includegraphics[width=0.14\linewidth]{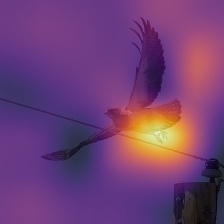}}}%
	\end{center}
	\caption{RISE with different baseline images.}
	\label{fig:rise_different_baselines}
\end{figure}

The systematic choice of a suitable baseline image is still an open research question \cite{dabkowski_gal_2017,fong_iccv_2017,Petsiuk2018rise,Sundararajan2017AxiomaticAF,zeiler_fergus_2014}. The proposed condition for the baseline image is, that it is uninformative \cite{Sundararajan2017AxiomaticAF} and therefore different baseline images have been presented in the literature \cite{samek_etal_2017}. The following gives an overview over the types of baseline images considered in our work. The constant baseline images are given by $\vec{A}_{\mathrm{\gamma}} = \gamma \cdot \vec{1}_{\mathrm{m \times n \times 3}}, \forall \gamma \in \Gamma \subseteq \{0, ..., 255\}$. Note that $\gamma = 0$, $\gamma = 127$, and $\gamma = 255$ refer to a black, gray and white baseline image, respectively. The zero baseline image after preprocessing, which is used by \cite{Petsiuk2018rise}, is $\vec{A}_{\mathrm{inv}} = g^{-1}(\vec{0}_{\mathrm{m \times n \times 3}})$. Applying the preprocessing function $g$ to $\vec{A}_{\mathrm{inv}}$, yields $g(\vec{A}_{\mathrm{inv}}) = \vec{0}_{\mathrm{m \times n \times 3}}$. The blurred baseline image is defined as $\vec{A}_{\sigma} = \pi(\vec{X}, \sigma), \forall \sigma \in \Sigma$, where $\pi$ applies a Gaussian blur with standard deviation $\sigma$ and $\Sigma$ is the set of standard deviations. In the past, values for the standard deviation of $\sigma = 3$ \cite{samek_etal_2017} and $\sigma = 5$ \cite{Petsiuk2018rise} have been chosen, however, applying these standard deviations can result in an output probability for the target class much greater than 0, which is not an uninformative baseline image. Therefore, we follow the choice of $\sigma = 10$ as in \cite{dabkowski_gal_2017,fong_iccv_2017}, which yields a baseline image with output probability of approximately 0. Note, that the blurred baseline image is the only baseline image which uses local information from the current input image $\vec{X}$.
We combine the above baseline images in a set of baseline images:
\begin{equation}
\mathcal{A}(\Gamma, \Sigma, g) =
\left\{
\vec{A}_{\mathrm{\gamma}}, 
\vec{A}_{\mathrm{inv}},
\vec{A}_{\sigma},
\vert \forall \gamma \in \Gamma, \forall \sigma \in \Sigma \right\}.
\end{equation}
Specifically the set  $\mathcal{A}_e = \mathcal{A}(\{0, 127, 255\}, \{10\}, g)$ is used in the following experiments. The set could also be extended to include a baseline image sampled from a uniform or Dirichlet distribution, the constant average from the current input, and a constant input image averaged over the dataset. However, the chosen set of baseline images is already sufficient for raising robustness issues for input perturbation methods. 
We adapt RISE to enable the application of different baseline images: 
\begin{equation}
\vec{S} = \frac{1}{\mathbb{E}(\mathcal{M}_N) \cdot N} \sum_{i=1}^N f^c(\Phi(\vec{X}, \vec{A}, q(\vec{M}_i)),
\end{equation}
where $\vec{A} \in \mathcal{A}_e$. Each saliency map is generated for examples $\vec{X} \in \mathcal{D}_{\mathrm{val}}$ for the class $c_{\mathrm{max}}$ with the highest output probability for that input image and $N = 16384$. This choice of $N$ is a compromise between accuracy and computational effort. If the $\mathrm{L}^2$-distance among saliency maps for an input image is greater than $d_{\mathrm{max}}$, the input will not be considered for this experiment. Therefore, we only take into account examples, which have converged.
An overview of four examples, where different baseline images yield different results, is given in Figure \ref{fig:rise_different_baselines}. The top row shows the tested input images and the two bottom rows each display the application of a different baseline image when creating the RISE saliency map for the specific input image. Depending on the baseline image, RISE attributes more or less importance to certain parts of the input. Comparing Figures \ref{fig:rise_different_baselines_gray_people} and \ref{fig:rise_different_baselines_black_people}, fewer or more people are highlighted based on the choice of baseline image. Similarly, in Figures \ref{fig:rise_different_baselines_white_bird} and \ref{fig:rise_different_baselines_black_bird} the bird's left wing and the bird's feet are highlighted, respectively. Depending on the selected baseline image the saliency map for RISE can completely change. RISE is therefore highly dependent on the choice of the baseline image. 
\subsection{Experiments on MoRF and LeRF}
MoRF and LeRF provide an objective and automatable metric for assessing different saliency methods. Here, we use MoRF to compare one saliency map generated by RISE with one generated by Grad-CAM \cite{grad_cam}. We limit the presented experiments to MoRF and refer the reader to the supplementary material \cite{supplementary_material} for experiments on LeRF, which yield the same propositions. RISE and Grad-CAM have been chosen for demonstration purposes only. The choice of different saliency methods for the comparison is equally reasonable. The generation of each saliency map uses the standard parameters presented in the original paper for Grad-CAM \cite{grad_cam}. We run MoRF with $r = 1$, such that the input images are covered pixel by pixel. The function $m: [0, 1] \times \left[0, 1\right]^{224 \times 224} \rightarrow \{0, 1\}^{224 \times 224}$ gives the mask induced by a saliency map $S$, where the relative occluded area of the input image is given by $\alpha$. 

In the first set of experiments we investigate the behavior of MoRF for different single input images. Similar as in \cite{Petsiuk2018rise}, we also only use the zero after preprocessing baseline image for MoRF and the blurred baseline image for LeRF. In the examples in Table \ref{tab:morf_same_baseline}, we find that comparing the AOC for MoRF for the single input images results in a conflicting assessment on which saliency method is preferable. For example, for the input image on the left side of Table \ref{tab:morf_same_baseline}, the AOC for the saliency map from Grad-CAM is greater than the AOC for the saliency map from RISE. However, for the input image on the right the AOC for the saliency map generated by RISE is greater. Note, that the output probability of the original input for the target class is displayed at $\alpha = 0$, when the image is not covered at all. According to the graphs in Table \ref{tab:morf_same_baseline} the selected baseline images at $m(1, \vec{S}) = \vec{A}$ also fulfill the requirement of being uninformative since the output probabilities are close to zero.
\begin{table*}[hpt!]
	\caption{MoRF with the zero after preprocessing baseline image for input images of the classes \texttt{chimpanzee} (left) and \texttt{beacon} (right).}
	\begin{center}
		\includegraphics{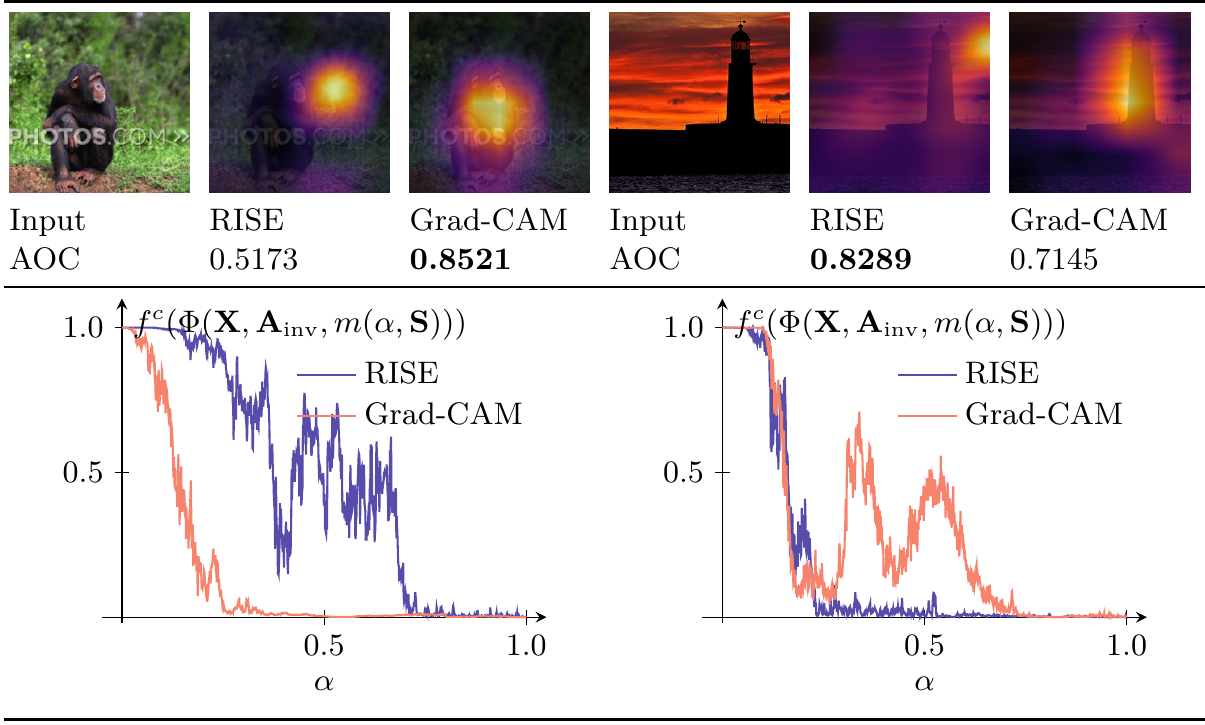}
	\end{center}
	\label{tab:morf_same_baseline}
\end{table*}
\begin{table*}[hpt!]
	\caption{MoRF with different baseline images for the class \texttt{flamingo}.}
	\begin{center}
		\includegraphics{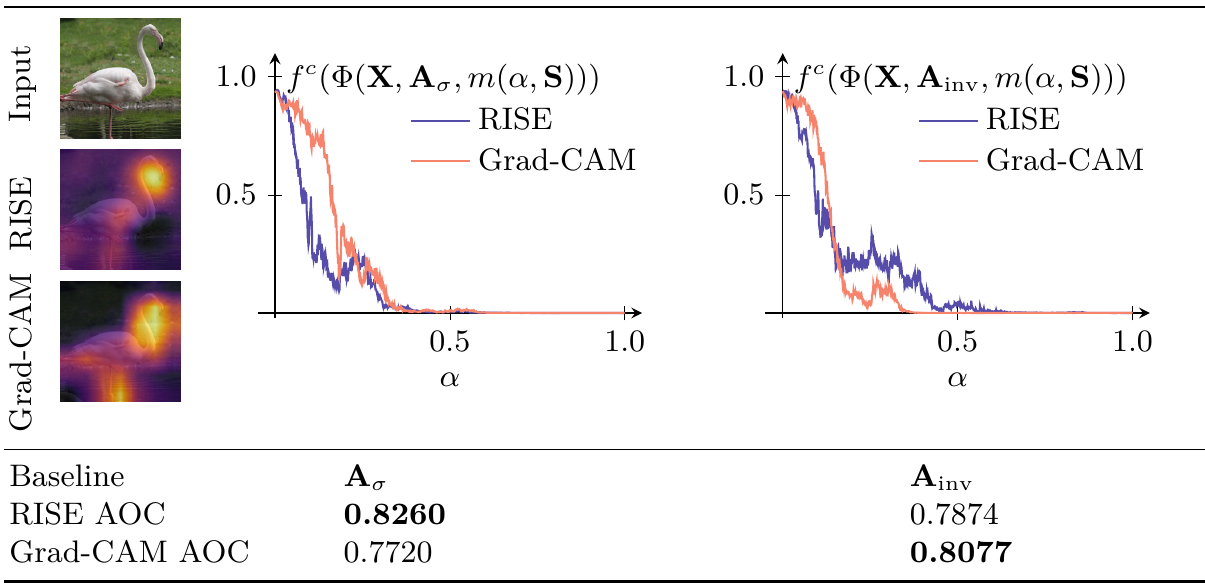}
	\end{center}
	\label{tab:morf}
\end{table*} 	

The subsequent set of experiments illustrate the dependency of the input perturbation methods on the choice of the baseline image. Again we use  MoRF to compare one saliency map generated by RISE with one generated by Grad-CAM. Here only the results for the $\vec{A}_{\sigma}$ and $\vec{A}_{\mathrm{inv}}$ baseline images are reported, because these are the baseline images used for the insertion and deletion metrics by \cite{Petsiuk2018rise}. However, the results below can also be observed when applying different baseline images. 
The results for MoRF on different input images are shown in Table \ref{tab:morf}. Each graph compares the saliency maps from RISE and Grad-CAM. The table shows that by only changing the baseline image, the result of MoRF inverts. While one saliency method is superior using a specific baseline image, it is inferior using another baseline image. Furthermore, RISE performs better than Grad-CAM with the $\vec{A}_{\sigma}$ baseline image for MoRF but worse than Grad-CAM using the same baseline image for LeRF. Therefore, the results from MoRF and LeRF are highly dependent on the choice of the baseline image. Note, that even though the RISE saliency map was created using the zero after preprocessing baseline image, the AOC for MoRF with the blurred baseline image $\vec{A}_{\sigma}$ is greater than the AOC for MoRF with the zero after preprocessing baseline image. This shows that running MoRF with the same baseline image, that was used to create the RISE saliency map, does not lead to a greater AOC for RISE.

\section{Discussion}
\label{sec:discussion}
RISE does not converge with the parameters presented by Petsiuk \etal \cite{Petsiuk2018rise} for every input image. Therefore, we implemented a convergence check with three independent sets of masks using the $\mathrm{L}^2$-distance. As shown by the experiments, different sets of parameters do not necessarily converge for all input images. Furthermore, the experiments show that varying the parameters for the mask generation can change the resulting saliency map significantly. In the case of Figure \ref{fig:rise_conv_goldfish} multiple goldfish are visible and RISE does not converge using the standard parameters. In this example, the probability of at least one goldfish being visible is very high when covering half the image ($p=0.5$). This results in a high prediction score for each masked input image. This could be one possible explanation, why the RISE saliency map is not able to converge in this case. The above example is more likely to converge if fewer goldfish are revealed with each single mask. In practice, a smaller value for $p$ led to convergence, if multiple instances of the same object are visible in the input image. While $w$ can impact convergence, we did not find indicators for a correlation between $w$ and the object size. In general, it is not clear how to decide on the parameters for running RISE on an input image. Determining the set of parameters requires multiple executions of RISE, which increases the computational effort. It is also not obvious how to select one saliency map for assessment, in the case that multiple sets of parameters result in a converged saliency map for an input image. The presented experiments on RISE also raise the issue of reproducibility of saliency maps with RISE.

The application of MoRF and LeRF on single input images when using the same baseline images, shows that the outputs for MoRF and LeRF do not agree for the saliency maps for the different input images in a dataset, which supports the recent findings in \cite{sanity_checks_for_metrics}. While, \cite{samek_etal_2017} average the AOC scores for MoRF or AUC scores for LeRF over the whole dataset, we want to highlight that MoRF and LeRF are not normalized. We argue that this puts input images with a low output probability for the target class at a disadvantage, since they generally exhibit a lower AOC for MoRF or a lower AUC for LeRF. For example, consider MoRF and two input images, which receive an output probability for the target class of 0.4 and 0.9, respectively. The maximum possible AOC for the first image is significantly lower than the maximum possible AOC for the second one.

The results from MoRF, LeRF, and RISE suggest, that these metrics and methods are highly dependent on the choice of the baseline image. Each baseline image introduces a bias into the input image.
Even an all zero input can yield activations for the ResNet-50 because of nonzero bias terms. The choice of a neutral baseline image is therefore nontrivial. Creating saliency maps from RISE with different baseline images can result in highlighting contradicting parts of the input image. Consequently, a specific baseline image can lead to an incorrect conclusion when assigning importance to certain elements in the input. Similarly, the experiments on the variation of baseline images display that by changing baseline images it is possible to invert the proposition from MoRF and LeRF on which saliency method is superior. 

\section{Conclusion}
\label{sec:conclusion}
In this paper we investigated input perturbation methods and metrics under changing parameters and baseline images. First, the experiments on RISE revealed, that convergence is not guaranteed for the majority of analyzed input images. Convergence could still be achieved for certain inputs, by varying the parameters for $w$, $h$ and $p$. However, different sets of parameters yield significantly distinct results, which might contradict each other by highlighting different parts of the input image. Determining the best parameters by multiple independent runs of RISE with different parameters and checking convergence for each input image increases the amount of required computations. Furthermore, these additional runs and variations still do not guarantee non-contradicting saliency maps from RISE. 
Second, varying the baseline image has led to notably different saliency maps for RISE and changed the AOC and AUC values for MoRF and LeRF, respectively, such that a reliable evaluation of saliency methods is challenging. 

Our results suggest that input perturbation methods are unreliable for understanding CNNs' predictions and the current available methods should not be used in practice. Since RISE, MoRF, and LeRF lack desirable robustness properties, future work will explore the development of robust input perturbation methods along with additional experiments for determining their robustness.

%
%
\bibliographystyle{splncs04}
\bibliography{references}

\end{document}